\documentclass[10pt,twocolumn,letterpaper]{article}

\usepackage{wacv}      % To produce the REVIEW version for the algorithms track
\usepackage{graphicx}
\usepackage{amsmath}
\usepackage{amssymb}
\usepackage{booktabs}
\usepackage{multicol}
\usepackage{multirow}
\usepackage{rotating}
\usepackage{caption}
\usepackage{tikz}
\usepackage{color, colortbl}
\usepackage{xcolor}
\usepackage{makecell}
\usepackage[most]{tcolorbox}
\usepackage[linesnumbered,ruled,vlined]{algorithm2e}
\usepackage[pagebackref,breaklinks,colorlinks]{hyperref}

\usepackage[capitalize]{cleveref}
\crefname{section}{Sec.}{Secs.}
\Crefname{section}{Section}{Sections}
\Crefname{table}{Table}{Tables}
\crefname{table}{Tab.}{Tabs.}

\newcommand\inline{\noindent\textbf}

\def\wacvPaperID{1762} % *** Enter the WACV Paper ID here
\def\confName{WACV}
\def\confYear{2024}

\SetKwInput{KwInput}{Input}                % Set the Input
\SetKwInput{KwOutput}{Output}              % set the Output

\begin{document}

\title{Top-Down Beats Bottom-Up in 3D Instance Segmentation}

\author{Maksim Kolodiazhnyi
\and Anna Vorontsova
\and Anton Konushin
\and Danila Rukhovich\\
Samsung Research\\
{\tt\small \{m.kolodiazhn, a.vorontsova, a.konushin, d.rukhovich\}@samsung.com} \\
}

\maketitle

%%%%%%%%% ABSTRACT
\begin{abstract}
   Most 3D instance segmentation methods exploit a bottom-up strategy, typically including resource-exhaustive post-processing. For point grouping, bottom-up methods rely on prior assumptions about the objects in the form of hyperparameters, which are domain-specific and need to be carefully tuned. On the contrary, we address 3D instance segmentation with a TD3D: the pioneering cluster-free, fully-convolutional and entirely data-driven approach trained in an end-to-end manner. This is the first top-down method outperforming bottom-up approaches in 3D domain. With its straightforward pipeline, it demonstrates outstanding accuracy and generalization ability on the standard indoor benchmarks: ScanNet v2, its extension ScanNet200, and S3DIS, as well as on the aerial STPLS3D dataset. Besides, our method is much faster on inference than the current state-of-the-art grouping-based approaches: our flagship modification is 1.9x faster than the most accurate bottom-up method, while being more accurate, and our faster modification shows state-of-the-art accuracy running at 2.6x speed. Code is available at \url{https://github.com/SamsungLabs/td3d}.
\end{abstract}

%%%%%%%%% BODY TEXT
\section{Introduction}
\label{sec:intro}

With the emergence of AR/VR, 3D indoor scanning, and household robotics, 3D instance segmentation becomes a key technology facilitating scene understanding. It is a holistic and challenging task of finding objects in 3D point clouds, predicting their semantic labels, and assigning an instance ID for each object. 

Two major 3D instance segmentation paradigms have been introduced so far~\cite{sun2022neuralbf}. \textit{Bottom-up} methods learn per-point embeddings and use them to cluster points so that they form a set of proposals. \textit{Top-down} directly predict instance proposals as object proxies, which are then filtered via non-maximum suppression, and refined via mask segmentation individually. 

In 2D instance segmentation, most state-of-the-art methods follow the top-down paradigm. Unfortunately, 2D methods that work well on a pixel grid cannot be directly adapted to process unstructured and sparse 3D points, and bottom-up methods dominate the field of 3D point cloud processing. Accordingly, the recent progress in 3D instance segmentation has been associated with improving components of bottom-up approaches: different ways of selecting points to be grouped have been studied~\cite{he2021dyco3d}, advanced feature aggregation strategies have been proposed~\cite{chen2021hierarchical, liang2021instance}, with estimates being refined via elaborate post-processing schemes~\cite{jiang2020pointgroup, vu2022softgroup}. In the meantime, top-down methods have been out of the spotlight. 

\begin{figure}[t!]
    \centering
    \includegraphics[scale=0.25]{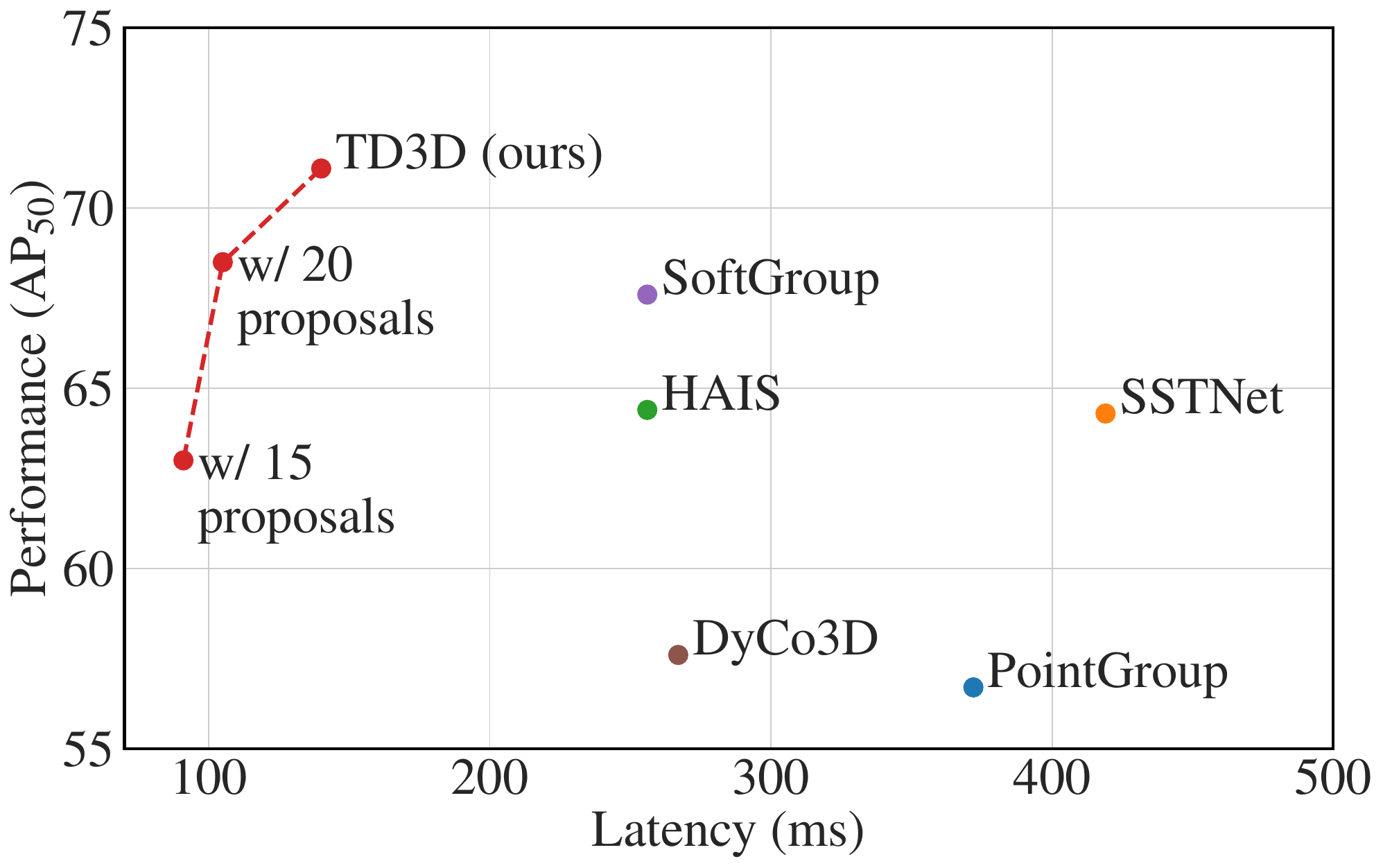}
    \caption{Prediction accuracy on ScanNet against latency. TD3D modifications (marked red) have a different number of proposals. Our top-performing default TD3D model surpasses existing methods in both detection accuracy and latency, while the faster modifications demonstrate an impressive inference speed with a comparable quality.}
    \label{fig:teaser}
\end{figure}

Nevertheless, bottom-up 3D instance segmentation methods have crucial drawbacks, limiting their performance. Besides being computationally-expensive, bottom-up approaches are sensitive to the values of numerous hyperparameters. Particularly, they might fail to find a proper balance between over-fragmented and accidentally merged masks and have limited generalization ability to complex scenes with objects of varying scales.

Our goal is to prove the top-down paradigm has great potential, which is yet unleashed in the 3D domain. In this paper, we tackle the challenging 3D instance segmentation task with TD3D, a top-down, fully-convolutional, simple approach trained end-to-end in a fully data-driven way. We conduct extensive experiments on the ScanNet v2, ScanNet200, S3DIS and STPLS3D datasets, and report competitive results for all these benchmarks.

Overall, our contributions are three-fold: \begin{itemize}
    \item We develop the world's first fully sparse convolutional cluster-free 3D instance segmentation approach, dubbed TD3D; 
    \item We introduce the first top-down method that supersedes bottom-up competitors, hence questioning the dominating paradigm in 3D instance segmentation; 
    \item We establish a state-of-the-art in both accuracy and speed: our flagship model is 1.9x faster than the best bottom-up approach, and we also show that state-of-the-art accuracy can be achieved with a 2.6x speed-up;
\end{itemize}

\begin{figure*}[t]
    \begin{center}
        \includegraphics[width=0.925\linewidth]{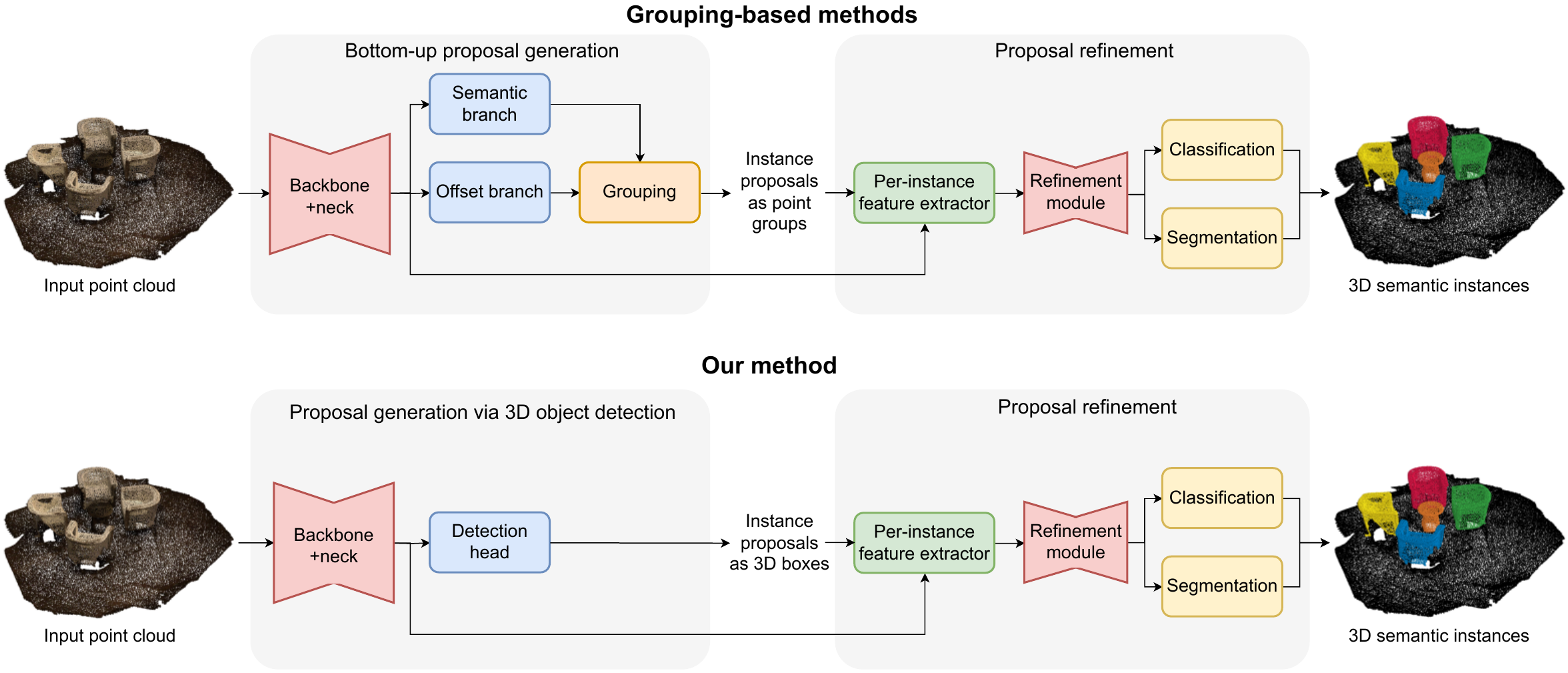}
    \end{center}
    \caption{Scheme of TD3D in comparison with state-of-the-art grouping-based methods (e.g., PointGroup~\cite{jiang2020pointgroup}, SoftGroup~\cite{vu2022softgroup}). Our proposal refinement is the same as in SoftGroup~\cite{vu2022softgroup}, while proposal generation significantly differs. Specifically, bottom-up proposal generation returns \textit{point groups}, while we leverage 3D object detection that outputs instance proposals defined by \textit{3D object bounding boxes} instead.}
\label{fig:scheme}
\end{figure*}

\section{Related Work}
\label{sec:related}

\inline{Bottom-up methods.} Up until very recently, grouping-based bottom-up methods have dominated the field. In SGPN~\cite{wang2018sgpn}, a similarity matrix for all 3D point pairs is learned, and the most similar points are assembled into instances. %MTML~\cite{lahoud2019mtml} proposed a multi-task learning strategy for grouping points. 
3D-SIS~\cite{hou2019sis} utilizes RGB images as an additional source of data and merges 3D features from a point cloud with backprojected 2D features extracted from RGB images.
ASIS~\cite{xinlong2019associatively} uses spatial discriminative loss to learn point-level embeddings and generates instance masks via a mean-shift algorithm.
3D-MPA~\cite{engelmann20203dmpa} predicts instance centers and refines initial instance proposals with a graph convolutional network.
Additional estimates are used to guide clustering, e.g., OccuSeg~\cite{han2020occuseg} predicts occupancy, while PointGroup~\cite{jiang2020pointgroup} assigns 3D points with semantic labels and center votes. In HAIS~\cite{chen2021hierarchical}, clustering is performed in a hierarchic manner. SSTNet~\cite{liang2021instance} aggregates point-wise semantic and instance-level features, using a semantic superpoint tree (SST) with superpoints as leaves. SoftGroup~\cite{vu2022softgroup} leverages a 3D sparse network to group 3D points according to the predicted soft semantic scores and refines the obtained proposals with a 3D U-Net-like network. DyCo3D~\cite{he2021dyco3d} also employs refinement, yet incorporates dynamic convolutions. 

\inline{Top-down methods.} Top-down 3D instance segmentation methods directly generate object proposals and then predict or refine masks for each proposal. 3D-BoNet~\cite{yang2019learning} applies Hungarian matching and outputs a fixed set of proposals in the form of non-oriented 3D bounding boxes. Instead of regressing 3D bounding boxes, GSPN~\cite{li2019gspn} employs an analysis-by-synthesis strategy to predict instance shapes. NeuralBF~\cite{sun2022neuralbf} generates the affinity of points in the point cloud to a query point and uses coordinate networks representing convex domains to model the spatial affinity in the neural bilateral filter.

\inline{3D object detection.} Modern 3D object detection methods can be categorized into voting-based, transformer-based, and 3D convolutional. Voting-based methods extract per-point features, merge them into an object proposal, and accumulate features of points within each group; overall, they use point grouping similar to bottom-up 3D instance segmentation methods. Instead of domain-specific heuristics and hyperparameters, transformer-based methods use end-to-end learning and forward pass on inference. Both voting- and transformer-based methods have scalability issues, making them impractical. Differently, top-down 3D convolutional methods represent point clouds as voxels, which makes them more memory-efficient and allows scaling to large scenes without sacrificing point density. Up until very recently, such methods lacked accuracy, yet the last advances in the field allowed developing fast, scalable, and accurate methods~\cite{rukhovich2022fcaf3d}.

\section{Proposed Method}
\label{sec:method}

\begin{figure}[h!]
    \centering
    \includegraphics[scale=0.55]{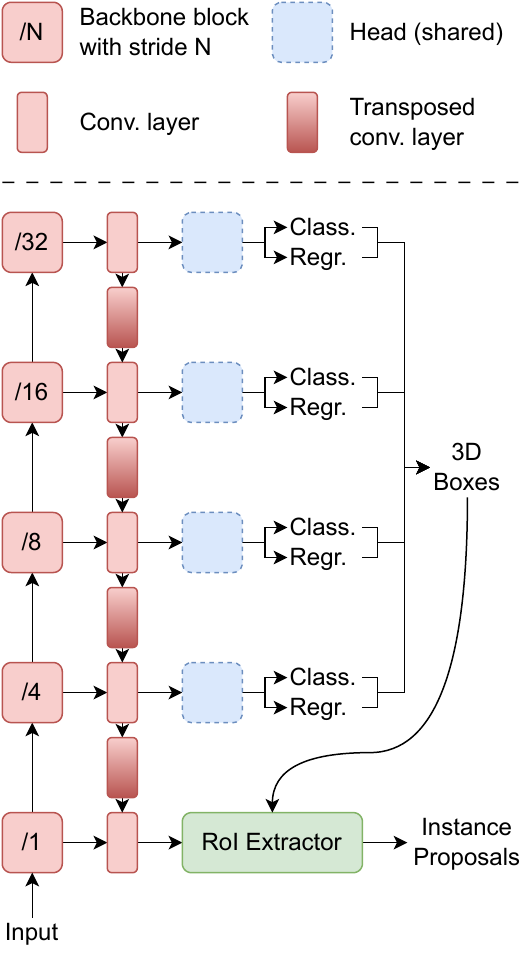}
    \caption{Our proposal generation scheme. 3D bounding boxes are estimated from the downsized 3D feature maps at higher network levels. Then, the predicted bounding boxes are used to select features from the 3D feature map of the original resolution.}
    \label{fig:scheme}
\end{figure}

The proposed method runs in two stages. First, it detects objects in a point cloud and extracts corresponding bounding boxes. These bounding boxes are interpreted as initial object proposals and then refined with a lightweight network to obtain final instance masks (Fig.~\ref{fig:scheme}). All operations within the pipeline are implemented through 3D sparse convolutions.  

\subsection{Proposal Generation}
\label{ssec:proposal-extraction}

\inline{3D Bounding Boxes.} We employ a 3D object detection method that outputs 3D object bounding boxes alongside object categories and confidence scores. A 3D object bounding box is parameterized as $(x, y, z, w, l, h)$, where $x, y, z$ denote the coordinates of the center of a bounding box, while $w, l, h$ are its width, length, and height, respectively.

Any conventional 3D object detection method can be employed for this purpose. We aim to avoid point grouping and follow a top-down paradigm at each stage of our pipeline, so we narrowed our search to the 3D convolutional-based methods. As the result, we opted for fast and efficient, fully-convolutional FCAF3D~\cite{rukhovich2022fcaf3d}.

The backbone in FCAF3D is a sparse ResNet~\cite{he2016deep} with sparse 3D convolutions. In the neck, the features on each level are processed with one sparse transposed 3D convolution and one sparse 3D convolution. To prevent sparsity growth, at most $N_{\text{vox}}$ voxels with the highest classification probabilities are selected at each level, where $N_{\text{vox}}$ equals the number of input points $N_{\text{pts}}$. The anchor-free FCAF3D head consists of two parallel sparse convolutional layers with weights shared across feature levels. For each location $(\hat{x}, \hat{y}, \hat{z})$, these layers output classification probabilities $\hat{\boldsymbol{p}}$ and bounding box regression parameters $\boldsymbol{\delta}$~\cite{rukhovich2022fcaf3d}.

\inline{Object Proposals.}
Given 3D bounding boxes, we extract the features from the corresponding regions of the 3D feature maps, which consists of voxels, where each voxel is defined by its own coordinate $(x, y, z)$ and feature vector $\Vec{f}$. We have five 3D feature maps of decreasing resolution: one map per feature level in the neck. At the first level, the 3D feature map has the same resolution as the input voxelized point cloud. The 3D feature maps at the second, third, fourth, and fifth level are 4x, 8x, 16x, and 32x smaller; they are used to estimate 3D bounding boxes. Then, these 3D bounding boxes and the first-level 3D feature map are processed with a RoI extractor, which selects voxels, whose centers are inside the given bounding box (Fig.~\ref{fig:scheme}). 

Eventually, these voxels with corresponding 3D features serve as initial object proposals. The pseudocode for RoI extraction is provided below (Algorithm 1).

\begin{algorithm}[!ht]
\DontPrintSemicolon
\LinesNumberedHidden
  
  \KwInput{
  
  Feature Map: $F = \{ v_i: (x_{v_i}, y_{v_i}, z_{v_i}, \Vec{f_i}) \ | \ i = 1 \ldots n\}$
  
  Bounding Boxes: $B = \{ (x_{b_j}, y_{b_j}, z_{b_j}, w_j, l_j, h_j) \ | \ j = 1 \ldots k \}$
  }
  \KwOutput{
  
  Proposals: $P = \{ P_t = \{v_0, v_1, \ldots v_{s_t} \} \ | \ t = 1 \ldots m, \ m \leq k \}$
  
  }

    \For{j=1 \ldots k}
    {
        $P_j := \emptyset $
    }
    \For{i=1 \ldots n}
    {
        \For{j=1 \ldots k} {
            $\ \ \delta x_1 = x_{v_i} - x_{b_j} + w_j / 2$
		$\delta x_2 = x_{b_j} - x_{v_i} + w_j / 2$
		$\delta y_1 = y_{v_i} - y_{b_j} + l_j / 2$
		$\delta y_2 = y_{b_j} - y_{v_i} + l_j / 2$
		$\delta z_1 = z_{v_i} - z_{b_j} + h_j / 2$
		$\delta z_2 = z_{b_j} - z_{v_i} + h_j / 2$
  
            \If{$\min(\delta x_1, \delta x_2, \delta y_1, \delta y_2, \delta z_1, \delta z_2) > 0$} {
                $P_j := P_j \bigcup \{v_i\}$
            }
        }
    }

    \For{j=1 \ldots k}
     {
      
        \If{$| P_j | < threshold$}
            {
                $P := P \text{\textbackslash} P_j$
            }
      }
\label{algo:roi-extractor}
\caption{RoI Extractor}
\end{algorithm}

\subsection{Proposal Refinement}
\label{ssec:proposal-refinement}

Our proposal refinement module takes voxels with features from RoI extractor as inputs and predicts final instance masks. For this purpose, we consider a 3D tiny U-Net network (a U-Net style network with few layers) solving a binary segmentation task, that classifies voxels into \textit{foreground} and \textit{background}. For each voxel, all points inside this voxel are assigned with the same \textit{foreground} or \textit{background} label predicted by U-Net, which gives final per-point instance masks.

\subsection{Training Procedure}
\label{ssec:training-procedure}

We train our method end-to-end, updating both proposal generation and proposal refinement models simultaneously. The total loss is a sum of two proposal generation losses $\mathcal{L}_{cls}$, $\mathcal{L}_{reg}$ and a proposal refinement loss $\mathcal{L}_{seg}$: $$\mathcal{L} = \mathcal{L}_{cls} + \mathcal{L}_{reg} + \mathcal{L}_{seg}$$

\inline{Proposal generation.} Our proposal generation model is inherited from FCAF3D~\cite{rukhovich2022fcaf3d}, accordingly, we follow the original training procedure and use focal loss $\mathcal{L}_{cls}$ and IoU loss $\mathcal{L}_{reg}$ to penalize classification and regression errors, respectively. During training, this model outputs 3D object bounding boxes parameterized with their centers and sizes (length, width, height), which we consider as initial object proposals.

\inline{Proposal refinement.} 
To train the proposal refinement model, it is essential to establish a correspondence between the proposals and the ground truth instances. This process has two stages. First, the centers of the 3D bounding boxes of the ground truth instances are calculated, and the ground truth instances are matched with the predicted 3D bounding boxes using the FCAF3D assigner~\cite{rukhovich2022fcaf3d}. For each ground truth bounding box, we select the last feature level where a 3D bounding box covers at least $N_{loc}$ voxels (if there is no such a feature level, the first feature level is chosen). Each voxel covered with a ground truth 3D bounding box is assigned with a semantic label and an index of this 3D bounding box. Respectively, the predicted 3D bounding box encoded with this voxel gets the same label and index.

At the second stage, the IoU assigner is employed. For each predicted 3D bounding box, we calculate IoU with all ground truth bounding boxes, and select the ground truth bounding box with the maximum IoU score. If FCAF3D and IoU assigners assigned the same ground truth 3D bounding box for the predicted 3D bounding box, then this assignment is considered trusted, and the predicted bounding box gets the label and index of the corresponding ground truth 3D bounding box. 

In~\cite{he2017mask}, the predicted 3D bounding boxes that do not have a corresponding ground truth 3D bounding box with an IoU score exceeding the given threshold, are filtered out. 
However, our ablation study reveals this strategy to be suboptimal (Tab.~\ref{tab:ablation-iou-assigner-thr}).

Finally, based on the predicted 3D bounding box, a proposal is extracted, and each voxel of the proposal is assigned with a semantic label and the index of the predicted 3D bounding box.

Our binary segmentation model is trained via minimizing $\mathcal{L}_{seg}$, which is calculated as a BCE loss between predicted and ground truth instance masks.

\section{Experiments}
\label{sec:experiments}

\subsection{Experimental Settings}
\label{ssec:setup}

% \begin{figure}[h!]
%     \centering
%     \setlength{\tabcolsep}{5pt}
%     %\resizebox{\linewidth}{!}{
%       \begin{tabular}{cc}
%         Ground truth & TD3D, ours \\
%         \includegraphics[width=0.45\linewidth]{images/scannet/scene0655_02_gt.png} & 
%         \includegraphics[width=0.45\linewidth]{images/scannet/scene0655_02_pred.png} \\
%         \includegraphics[width=0.4\linewidth]{images/scannet/scene0549_01_gt.png} &
%         \includegraphics[width=0.4\linewidth]{images/scannet/scene0549_01_pred.png} \\
%         \includegraphics[width=0.4\linewidth]{images/scannet/scene0412_01_gt.png} &
%         \includegraphics[width=0.4\linewidth]{images/scannet/scene0412_01_pred.png} \\
%     \end{tabular}
%     %}
%     \caption{Results of 3D instance segmentation of point clouds from ScanNet v2.}
%     \label{fig:qualitative-scannet}
% \end{figure}

\begin{figure*}[ht!]
    \begin{minipage}{.675\textwidth}
        \centering
        \setlength{\tabcolsep}{2pt}{
        \resizebox{\textwidth}{!}{
        \begin{tabular}{lllccccccc}
        \toprule
        \multirow{2}{*}{Paradigm} & \multirow{2}{*}{Method} & \multirow{2}{*}{Conference} & \multicolumn{3}{c}{Validation} & \multicolumn{3}{c}{Test} & Runtime \\
        \cmidrule(lr){4-6} \cmidrule(lr){7-9}
        & & & AP & AP\textsubscript{50} & AP\textsubscript{25} & AP & AP\textsubscript{50} & AP\textsubscript{25} & (in sec) \\
        \midrule
        \multirow{5}{*}{Bottom-up}  & PointGroup~\cite{jiang2020pointgroup} & CVPR'20   & 34.8      & 56.7      & 71.3      & 40.7      & 63.6      & 77.8      & 0.372 \\
                                    & SSTNet~\cite{liang2021instance}       & ICCV'21   & \textbf{49.4} & 64.3      & 74.0      & \textbf{50.6} & 69.8      & 78.9      & 0.419 \\
                                    & HAIS~\cite{chen2021hierarchical}      & ICCV'21   & 43.5      & 64.4      & 75.6      & 45.7      & 69.9      & 80.3      & 0.256 \\ 
                                    & DyCo3D~\cite{he2021dyco3d}            & CVPR'21   & 35.4      & 57.6      & 72.9      & 39.5      & 64.1      & 76.1      & 0.267 \\
                                    & SoftGroup~\cite{vu2022softgroup}      & CVPR'22   & 45.8      & \underline{67.6} & \underline{78.9} & \underline{50.4} & \textbf{76.1} & \underline{86.5} & 0.266 \\
        \midrule
        \multirow{5}{*}{Top-down}   & 3D-SIS~\cite{hou2019sis}              & CVPR'19   & -         & 18.7      & 35.7      & 16.1      & 38.2      & 55.8      & \textgreater10 \\
                                    & GSPN~\cite{li2019gspn}                & CVPR'19   & 19.3      & 37.8      & 53.4      & -         & 30.6      & -         & \textgreater10 \\
                                    & 3D-BoNet~\cite{yang2019learning}      & NeurIPS'19& -         & -         & -         & 25.3      & 48.8      & 68.7      & 9.174 \\
                                    & NeuralBF~\cite{sun2022neuralbf}       & WACV'23   & 36.0      & 55.5      & 71.1      & 35.3      & 55.5      & 71.8      & - \\
                                    & \textbf{TD3D (ours)}                  &           & \underline{47.3} & \textbf{71.2} & \textbf{81.9} & 48.9      & \underline{75.1} & \textbf{87.5} & \textbf{0.140}
                                    \\
        \bottomrule
        \end{tabular}
        }
        }
        \captionsetup{labelformat=empty}
        \captionof{table}{\textbf{Table 1}: Results on ScanNet v2. The best results are \textbf{bold}, the second best are \underline{underlined}. The runtime is measured using a single NVidia 3090 GPU. Our approach outperforms the previous state-of-art SoftGroup~\cite{vu2022softgroup} on the validation subset, while being 1.9 times faster.}
        \label{tab:results-scannet}
    \end{minipage}
    %\vfill
    \begin{minipage}{.275\textwidth}
        \vspace*{-0.75cm}
        \centering
        \setlength{\tabcolsep}{2pt}{
        \resizebox{\textwidth}{!}{
        \begin{tabular}{lccc}
        \toprule
        \multirow{2}{*}{Method} & \multicolumn{3}{c}{AP} \\
        \cmidrule(lr){2-4}
        & head & common & tail \\
        \midrule
        CSC~\cite{hou2021exploring}             & 22.3      & 8.2       & 4.6 \\
        Mink34D~\cite{choy20194dspatio}         & 24.6      & 8.3       & 4.3 \\
        LGround~\cite{rozenberszki2022language} & 27.5      & 10.8      & 6.0 \\
        \textbf{TD3D (ours)}                    & \textbf{33.2} & \textbf{17.7} & \textbf{10.3} \\
        \bottomrule
        \end{tabular}
        }
        }
        \captionsetup{labelformat=empty}
        \captionof{table}{\textbf{Table 2}: Results on the ScanNet200 test split. AP scores for the most frequent (\textit{head} of distribution), common, and rare (\textit{tail}) object categories are provided. The best results are \textbf{bold}. TD3D achieves 1.7x improvement of the previous state-of-the-art AP scores for \textit{common} and \textit{tail} categories.}
        \label{tab:results-scannet200}
    \end{minipage}
    %\vspace*{0.5cm}
\end{figure*}

\inline{Datasets.} The experiments are conducted on ScanNet v2~\cite{dai2017scannet}, ScanNet200~\cite{rozenberszki2022language}, S3DIS~\cite{armeni2016s3dis}, and recently introduced STPLS3D~\cite{chen2022stpls3d}. 
ScanNet v2~\cite{dai2017scannet} contains 1613 scans divided into training, validation, and testing splits of 1201, 312, and 100 scans, respectively. 3D instance segmentation is typically evaluated using 18 object classes. We report results on both validation and hidden test splits. ScanNet200~\cite{rozenberszki2022language} extends the original ScanNet semantic annotation with fine-grained categories with the long-tail distribution. The training, validation, and testing splits are similar to the original ScanNet v2 dataset. The S3DIS dataset~\cite{armeni2016s3dis} features 272 scenes within 6 large areas. Following the standard evaluation protocol, we assess the segmentation quality on scans from Area 5, and via 6 cross-fold validation, using 13 semantic categories in both settings. STPLS3D~\cite{chen2022stpls3d} is a synthetic outdoor dataset emulating aerial photogrammetry. It covers 25 urban scenes of 6 km\textsuperscript{2}, densely annotated with 14 categories. We use the splits proposed in the original work~\cite{chen2022stpls3d}.

\inline{Metrics.} We use the average precision as a major metric. AP\textsubscript{50} and AP\textsubscript{25} are the scores obtained with IoU thresholds of 50\% and 25\%, respectively. AP is an average score with IoU threshold varying from 50\% to 95\% with a step of 5\%. %For S3DIS, we also report mean coverage (mCov), mean weighed coverage (mWCov), mean precision (mPrec), and mean recall (mRec).

\inline{Implementation details.} 
Our models are implemented using mmdetection3d framework~\cite{mmdet3d2020} based on Pytorch. We use MinkUNet14B as a binary segmentation model at the refinement stage. We train for 330 epochs on a single NVidia 3090 GPU with the Adam optimizer. The batch size is 4, and the initial learning rate is set to 0.001 and is reduced by 10 times after 280 and 320 epochs. Other implementation details are similar to FCAF3D~\cite{rukhovich2022fcaf3d}.
\subsection{Comparison to Prior Work}
\label{ssec:comparison}

\inline{ScanNet v2.} Results for validation and test splits of ScanNet v2 are presented in Tab.~\ref{tab:results-scannet}. Overall, TD3D is on par with previous state-of-the-art SoftGroup~\cite{vu2022softgroup} on the test split and shows superior results on validation. Another advantage of TD3D is its inference speed: according to the reported runtime, it is more than $1.8$x faster than any method that performs grouping.

\inline{ScanNet200.} We evaluate TD3D on the test split of ScanNet200 and report metrics in Tab.~\ref{tab:results-scannet200}. For either frequent, common, or rare categories, our method demonstrates a solid superiority over the existing approaches. The gain is especially tangible for less frequent categories, where TD3D improves previous state-of-the-art metrics by approximately $1.7$x times ($+6.9$ and $+4.3$ AP for \textit{common} and \textit{tail}, respectively).

\inline{S3DIS.} According to the Tab.~\ref{tab:results-s3dis}, TD3D surpasses other methods by at least $+5.9$ AP and $+8.9$ Prec\textsubscript{50} for Area 5 and $+2.1$ AP and $+2.8$ Prec\textsubscript{50} on 6-fold cross-validation. Meanwhile, if being pre-trained on ScanNet and fine-tuned on S3DIS, as proposed in~\cite{chen2021hierarchical, vu2022softgroup}, it consistently outperforms the previous state-of-the-art SoftGroup~\cite{vu2022softgroup} in both testing scenarios and in terms of all metrics (Tab.~\ref{tab:results-pretraining}).

\begin{table}[h!]
\centering
\setlength{\tabcolsep}{1pt}{
\resizebox{1\linewidth}{!}{
\begin{tabular}{lcccccccc}
\toprule
\multirow{2}{*}{Method} & \multicolumn{4}{c}{Area 5} & \multicolumn{4}{c}{6-fold CV} \\
\cmidrule(lr){2-5} \cmidrule(lr){6-9}
& AP & AP\textsubscript{50} & Prec\textsubscript{50} & Rec\textsubscript{50} & AP & AP\textsubscript{50} & Prec\textsubscript{50} & Rec\textsubscript{50} \\
\midrule
SGPN~\cite{wang2018sgpn}                & -         & -         & 36.0      & 28.7      & -         & -         & 38.2      & 31.2 \\
ASIS~\cite{xinlong2019associatively}    & -         & -         & 55.3      & 42.4      & -         & -         & 63.6      & 47.5 \\
3D-BoNet~\cite{yang2019learning}        & -         & -         & 57.5      & 40.2      & -         & -         & 65.6      & 47.6 \\
OccuSeg~\cite{han2020occuseg}           & -         & -         & -         & -         & -         & -         & 72.8      & 60.3 \\
3D-MPA~\cite{engelmann20203dmpa}        & -         & -         & 63.1      & 58.0      & -         & -         & 66.7      & 64.1 \\
PointGroup~\cite{jiang2020pointgroup}   & -         & 57.8      & 61.9      & 62.1      & -         & 64.0      & 69.6      & 69.2 \\
DyCo3D~\cite{he2021dyco3d}              & -         & -         & 64.3      & 64.2      & -         & -         & -         & - \\
MaskGroup~\cite{min2022maskgroup}      & -         & \underline{65.0} & 62.9      & \underline{64.7}      & -         & \textbf{69.9}      & 66.6      & 69.2 \\
SSTNet~\cite{liang2021instance}         & \underline{42.7}      & 59.3      & \underline{65.5}      & 64.2      & \underline{54.1}      & 67.8      & \underline{73.5}      & \underline{73.4} \\
\textbf{TD3D (ours)}                    & \textbf{48.6} & \textbf{65.1} & \textbf{74.4} & \textbf{64.8} & \textbf{56.2} & \underline{68.2} & \textbf{76.3} & \textbf{74.0} \\
\bottomrule
\end{tabular}
}
}
\caption{Results on S3DIS. The best results are \textbf{bold}, the second best are \underline{underlined}. Being superior in all metrics in both testing scenarios, our approach sets a new state-of-art in 3D instance segmentation.}
\label{tab:results-s3dis}
\end{table}

\begin{table}[h!]
    \centering
    \small
    \setlength{\tabcolsep}{2pt}{
    %\resizebox{1\linewidth}{!}{
    \begin{tabular}{lcccccc}
    \toprule
    \multirow{2}{*}{Method} & \multicolumn{3}{c}{Area 5} & \multicolumn{3}{c}{6-fold CV} \\
    \cmidrule(lr){2-4} \cmidrule(lr){5-7}
    & AP & Prec\textsubscript{50} & Rec\textsubscript{50} & AP & Prec\textsubscript{50} & Rec\textsubscript{50} \\
    \midrule
    HAIS~\cite{chen2021hierarchical} & -         & 71.1      & 65.0      & -        & 73.2      & 69.4 \\
    SoftGroup~\cite{vu2022softgroup} & 51.6 & 73.6 & 66.6 & 54.4 & 75.3 & 69.8 \\
    \textbf{TD3D (ours)}                 & \textbf{52.1} & \textbf{75.2} & \textbf{68.7} & \textbf{58.1} & \textbf{82.8} & \textbf{71.6} \\
    \bottomrule
    \end{tabular}
    %}
    }
    \caption{Results on S3DIS with the ScanNet v2 pre-training. The best results are \textbf{bold}. TD3D shows a solid improvement over SoftGroup~\cite{vu2022softgroup} in terms of all metrics.}
    \label{tab:results-pretraining}
\end{table}

\inline{STPLS3D.} We evaluate TD3D behind the indoor domain, scoring unexpectedly high on STPLS3D. In Tab.~\ref{tab:results-stpls3d}, we compare our approach against strong baselines: evidently, TD3D sets a new state-of-art, superseding SoftGroup by impressive +8.1 AP and +8.0 AP\textsubscript{50}.

\begin{figure*}
    \centering
      \begin{tabular}{cccc}
         & ScanNet & S3DIS & STPLS3D \\
        \rotatebox[origin=c]{90}{Input} &
        \makecell{\includegraphics[width=0.27\linewidth]{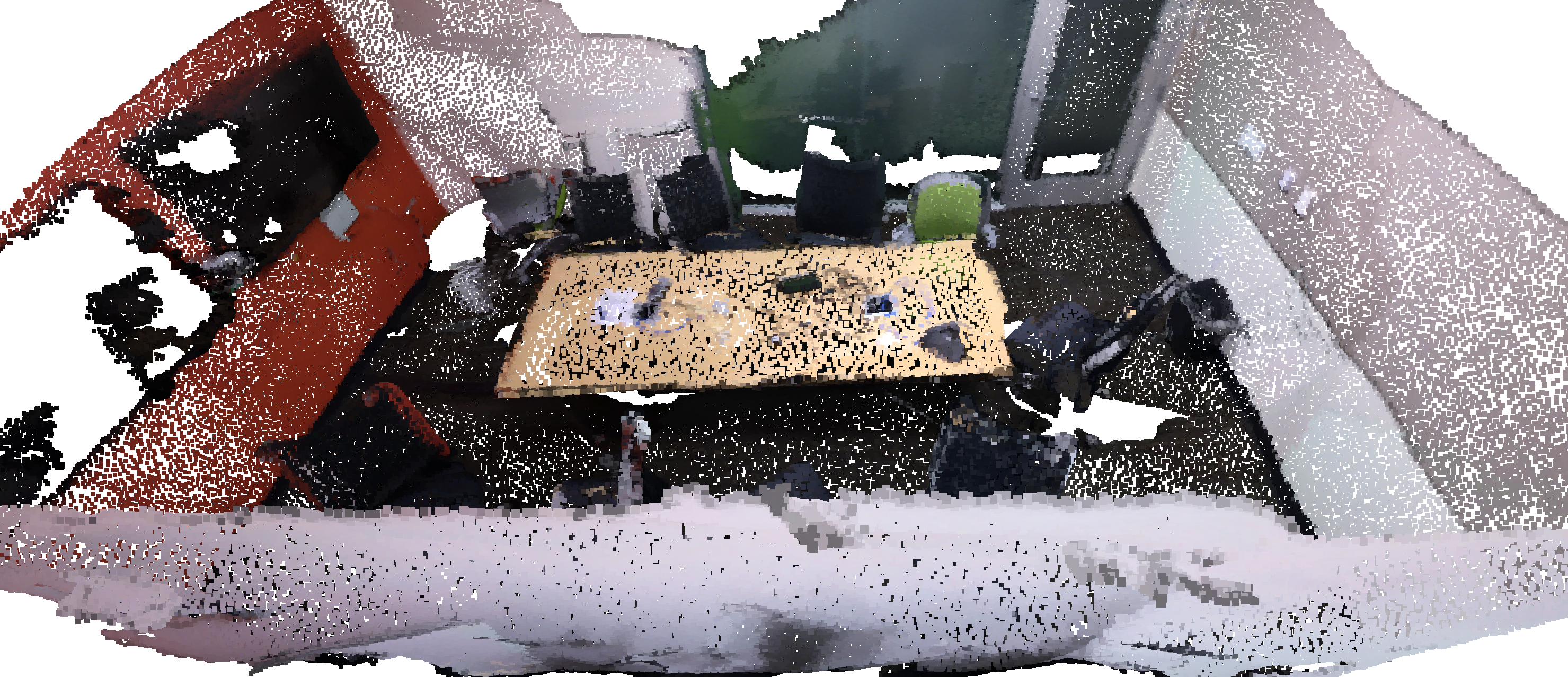}} &
        \makecell{\includegraphics[width=0.27\linewidth]{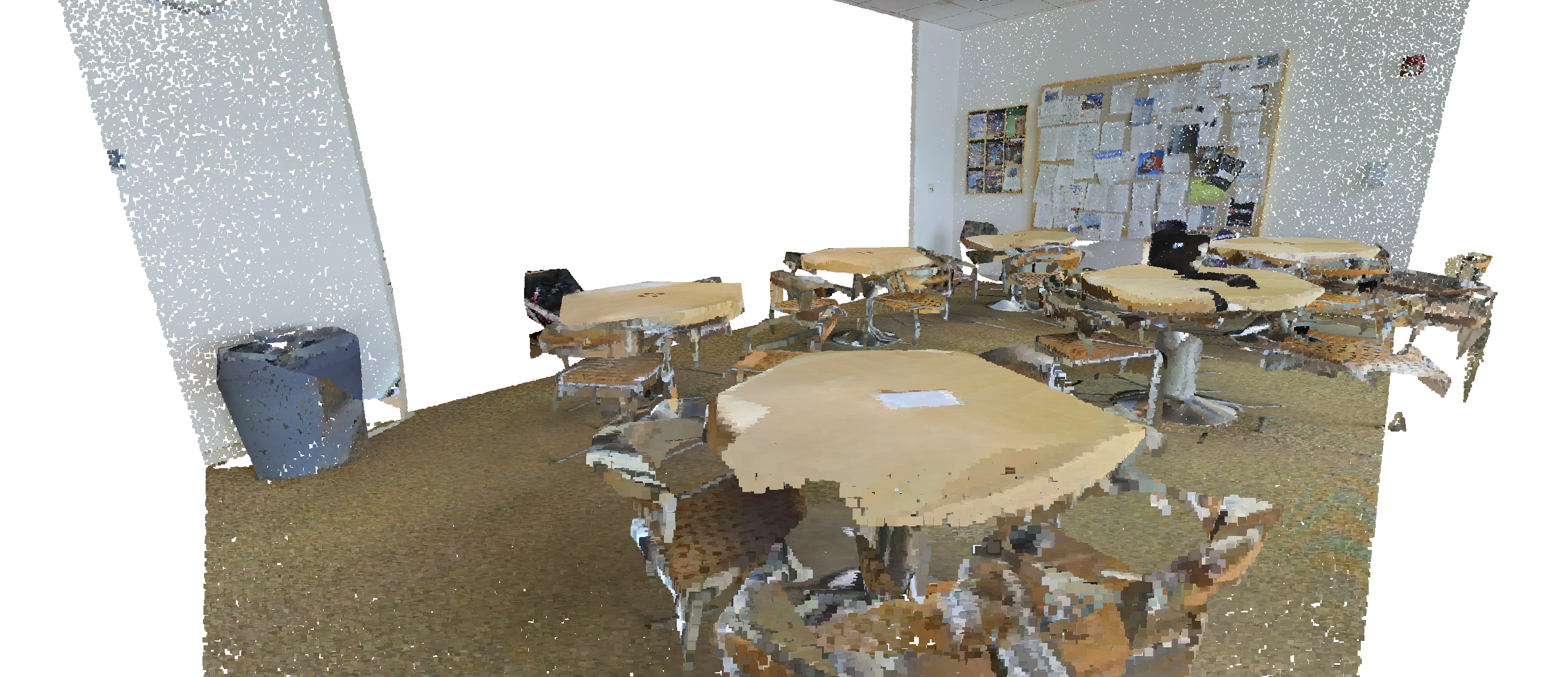}} &
        \makecell{\includegraphics[width=0.27\linewidth]{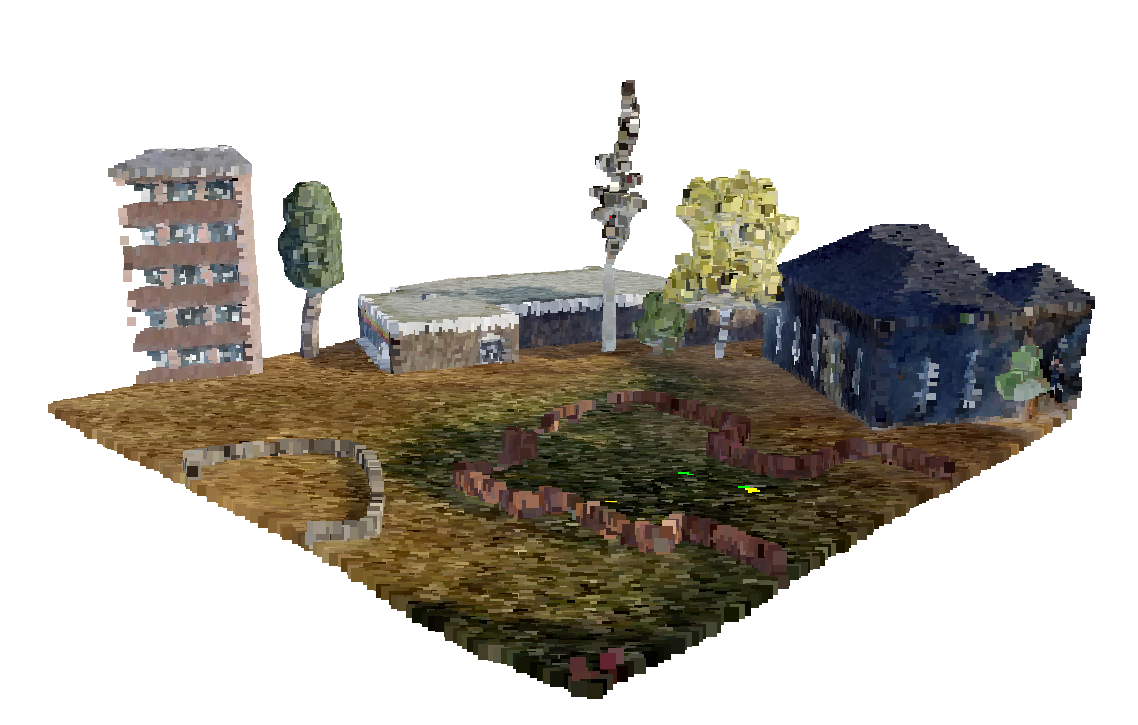}} \\
        \rotatebox[origin=c]{90}{Ground Truth} &
        \makecell{\includegraphics[width=0.27\linewidth]{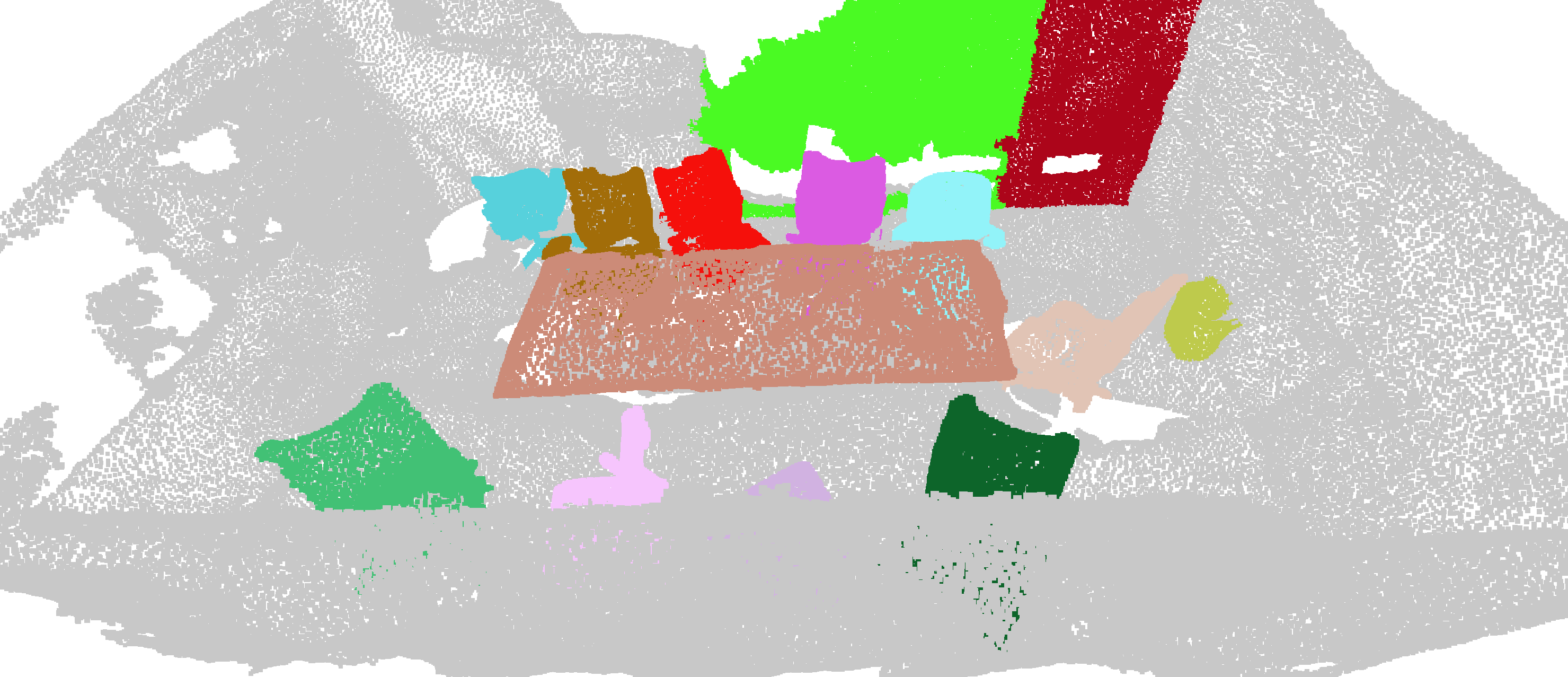}} &
        \makecell{\includegraphics[width=0.27\linewidth]{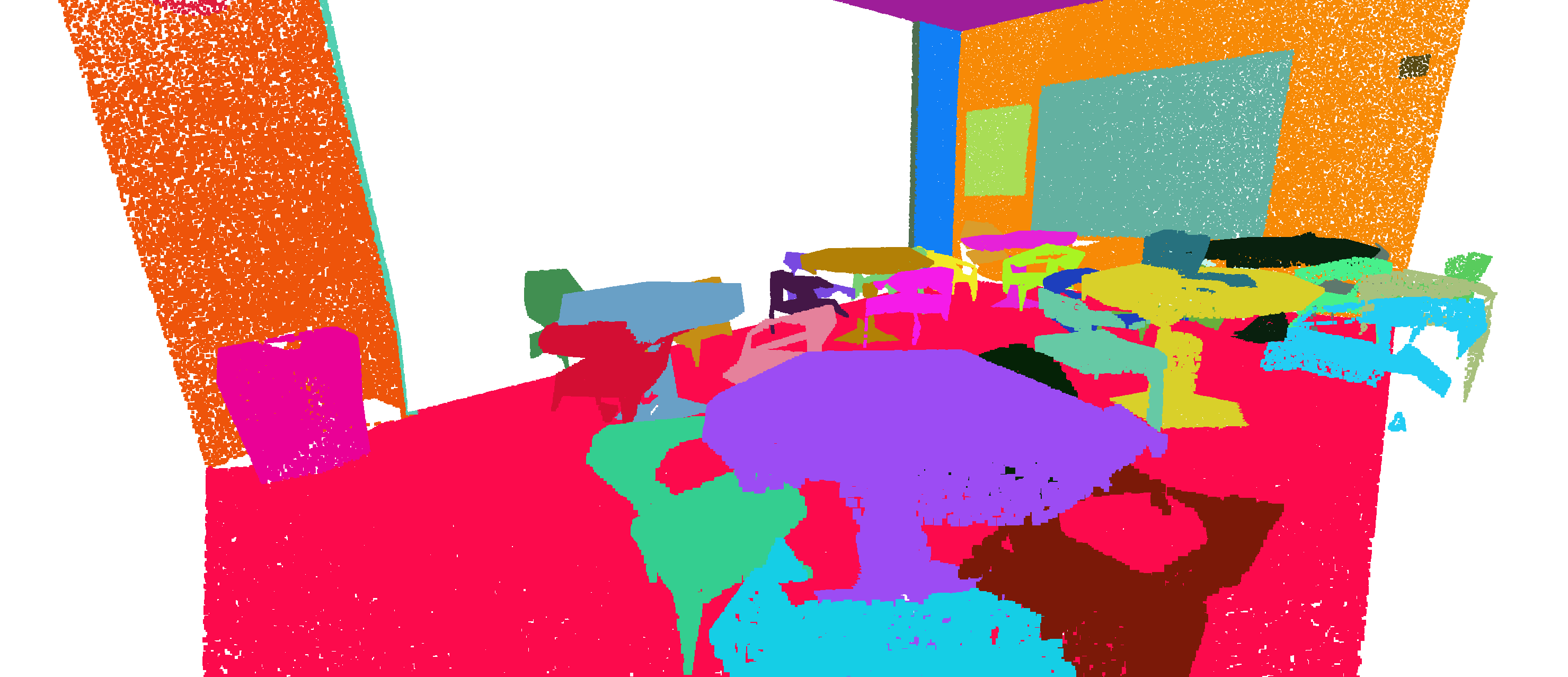}} &
        \makecell{\includegraphics[width=0.27\linewidth]{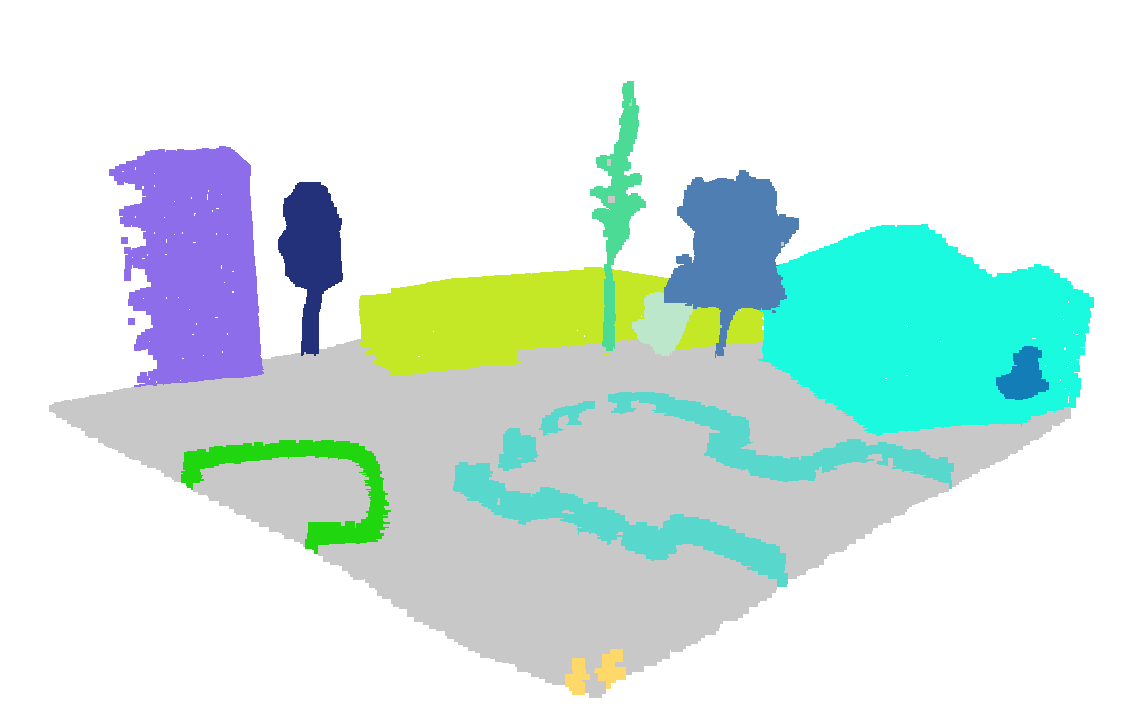}} \\
        \rotatebox[origin=c]{90}{Prediction} &
        \makecell{\includegraphics[width=0.27\linewidth]{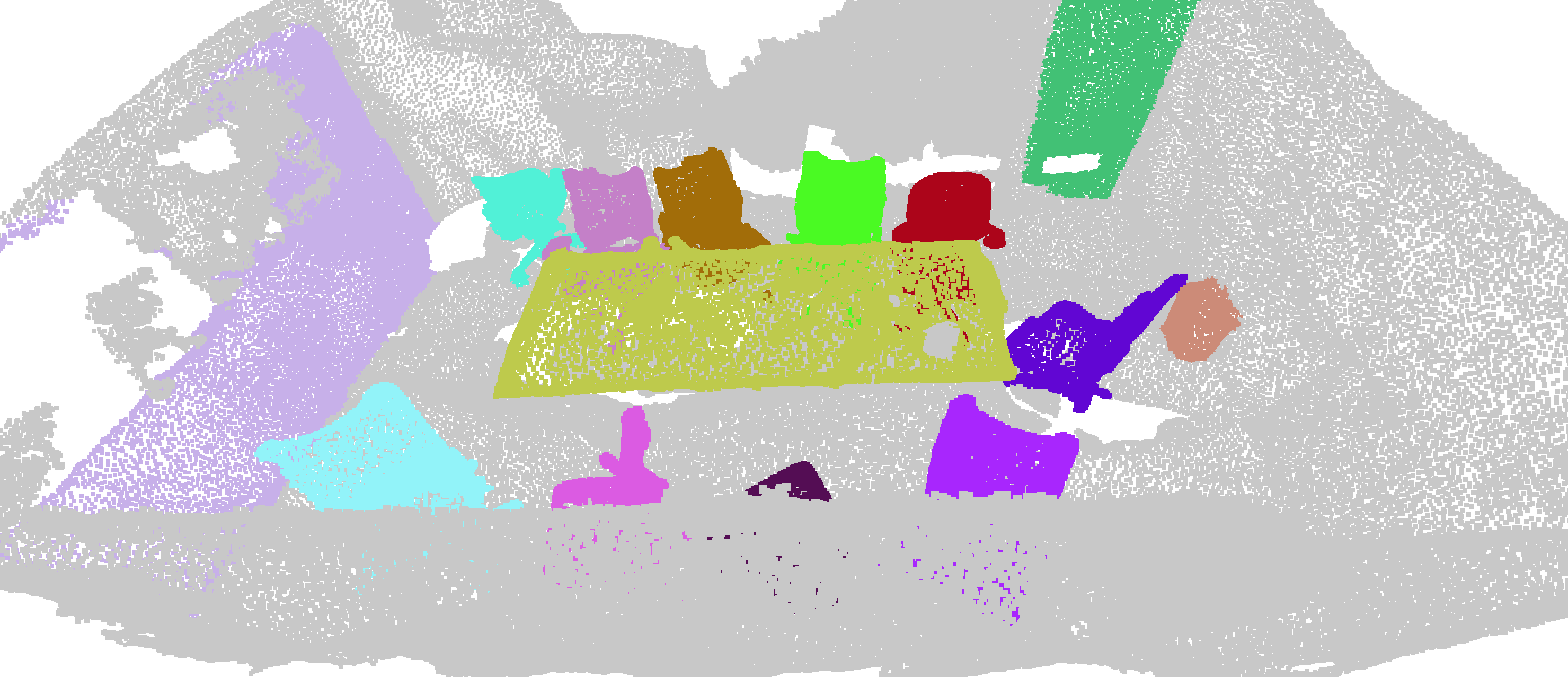}} &
        \makecell{\includegraphics[width=0.27\linewidth]{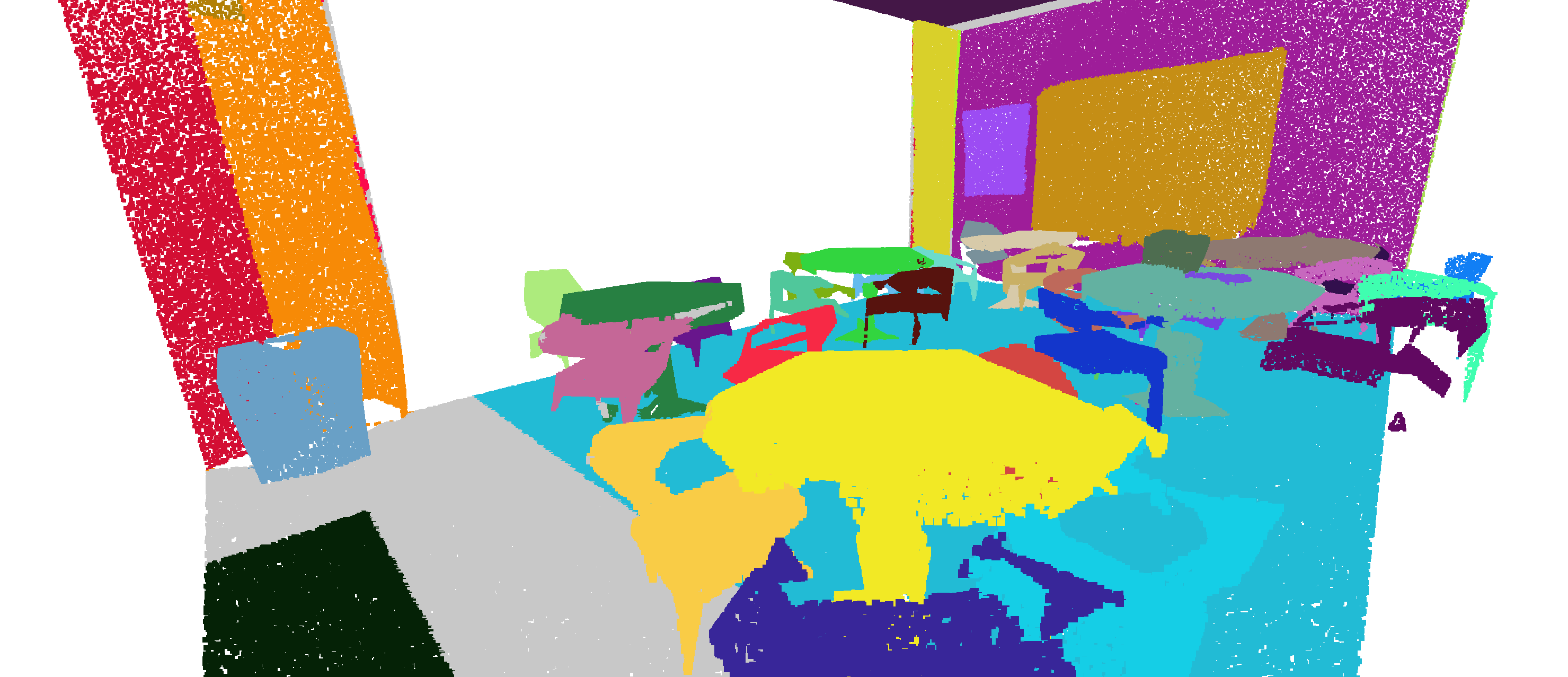}} &
        \makecell{\includegraphics[width=0.27\linewidth]{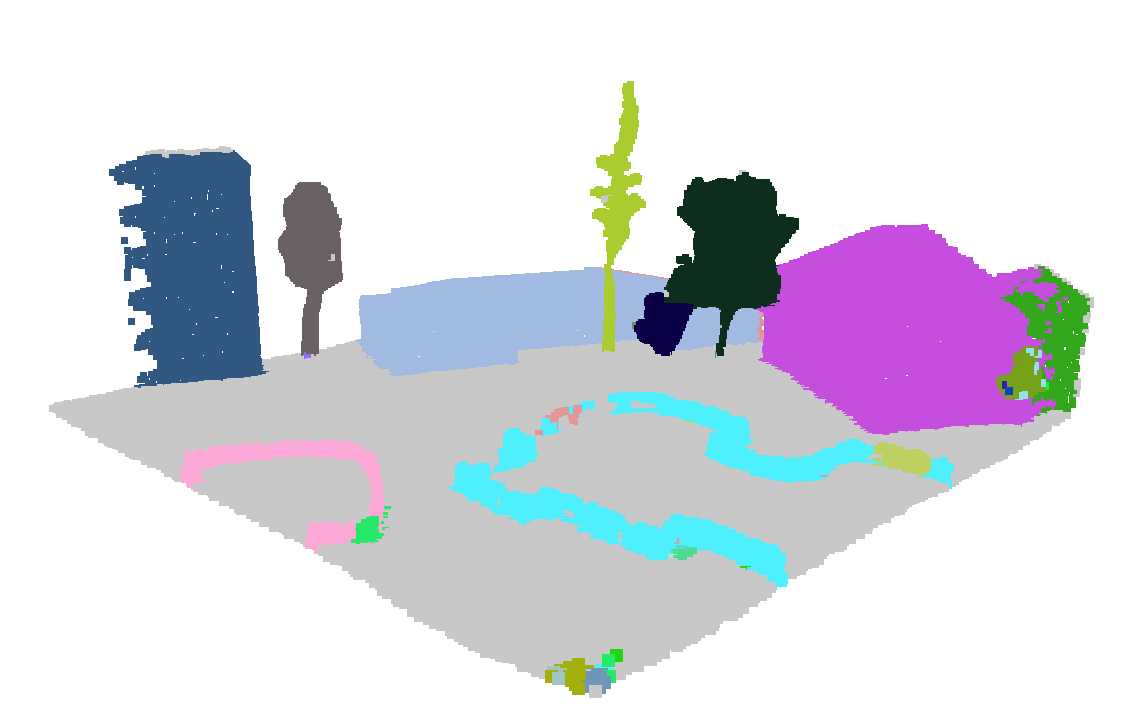}}
    \end{tabular}
    \caption{Results of 3D instance segmentation of point clouds from ScanNet, S3DIS, and STPLS3D.The first row is the original point cloud, the second row is the ground truth, the third row is the model predictions.}
    \label{fig:three-datasets}
\end{figure*}

\begin{table*}[h!]
\centering
\begin{tabular}{llccc}
\toprule
Method                      & Component             & Device  & Component time (ms) & Total (ms)           \\
\midrule
\multirow{3}{*}{PointGroup~\cite{jiang2020pointgroup}} & Backbone              & GPU     & 48                  & \multirow{3}{*}{372} \\
                            & Grouping              & GPU+CPU & 218                 &                      \\
                            & ScoreNet              & GPU     & 106                 &                      \\
\midrule
\multirow{3}{*}{HAIS~\cite{chen2021hierarchical}}       & Backbone              & GPU     & 50                  & \multirow{3}{*}{256} \\
                            & Hierarchical aggregation           & GPU+CPU & 116                 &                      \\
                            & Intra-instance refinement & GPU     & 90                  &                      \\
\midrule
\multirow{3}{*}{SoftGroup~\cite{vu2022softgroup}}  & Backbone              & GPU     & 48                  & \multirow{3}{*}{266} \\
                            & Soft grouping         & GPU+CPU & 121                 &                      \\
                            & Top-down refinement   & GPU     & 97                  &                      \\
\midrule
\multirow{4}{*}{SSTNet~\cite{liang2021instance}}     & Superpoint extraction & CPU     & 179                 & \multirow{4}{*}{419} \\
                            & Backbone              & GPU     & 34                  &                      \\
                            & Tree Network          & GPU+CPU & 148                 &                      \\
                            & ScoreNet              & GPU     & 58                  &                      \\
\midrule
\multirow{4}{*}{\textbf{TD3D, ours}} &  Backbone    & GPU     & 39                  & \multirow{4}{*}{\textbf{140}}    \\
                            & Proposal generation   & GPU     & 8                   &                      \\
                            & RoI extraction        & GPU     & 16                  &                      \\
                            & Proposal refinement   & GPU     & 77                  &         \\  
\bottomrule
\end{tabular} 

\caption{The inference time of TD3D and existing 3D instance segmentation methods, measured component-wise. All intermediate operations are performed on GPU, which allows achieving x1.8 speed-up in comparison with the fastest competitor, HAIS~\cite{chen2021hierarchical} and x1.9 speed-up with the most accurate competitor, SoftGroup~\cite{vu2022softgroup}.}
\label{tab:profiler}
\end{table*}

\begin{table}[h!]
\centering
%\small
\begin{tabular}{lll}
\toprule
Method     & AP  & AP\textsubscript{50} \\
\midrule
PointGroup & 23.3 & 38.5  \\
HAIS       & 35.1 & 46.7  \\
SoftGroup  & 46.2 & 61.8  \\
\textbf{TD3D, ours}       & \textbf{54.3} & \textbf{69.8} \\
\bottomrule
\end{tabular}
\caption{Results on STPLS3D. The proposed approach outperforms the competitors by a large margin.}
\label{tab:results-stpls3d}
\end{table}

\subsection{Qualitative Results}
\label{ssec:qualitative}

The original and segmented point clouds from ScanNet, S3DIS and STPLS3D datasets are depicted in Fig.~\ref{fig:three-datasets}.

\subsection{Performance}
\label{ssec:performance}

To provide an in-depth performance evaluation, we run a profiler to measure the time required to complete each component of our method: extracting 3D features with a 3D CNN, proposal generation, RoI extraction, and proposal refinement. We decompose several competing approaches into components similarly, and report the inference time component-wise in Tab.~\ref{tab:profiler}. Contrary to other listed methods, TD3D follows a top-down paradigm, which allows running all operations on GPU in an end-to-end pipeline. As the result, our approach is notably faster than the previous fastest method, HAIS~\cite{chen2021hierarchical}.

\subsection{Ablation Studies}
\label{ssec:ablation}

In this section, we analyze different components of our approach and measure the contribution to the final quality of each component. We do not introduce any changes into the 3D object detection part, but focus on the components that constitute the novelty of our approach: proposal generation and refinement. Namely, we investigate such aspects of proposal generation as the number of initial proposals and the RoI extraction threshold, and study the impact of the number of feature levels and the assigners used in the proposal refinement network. The ablation experiments are conducted on the ScanNet v2 validation set, following the same evaluation protocol as for the qualitative comparison.

\inline{Number of feature levels in the proposal refinement network.} We study how the size of the proposal refinement network affects the segmentation accuracy. Starting from 0 levels (all points in an initial proposal are included in the instance mask), and using no more than 4 levels (as in the backbone), we select the best option in terms of AP. As can be seen from Tab.~\ref{tab:ablation-unet}, AP grows with the number of levels; yet, we do not want our refinement model to be large, so we opt for four levels in the default version.

\begin{table}[h!]
    \centering
    %\small
    \begin{tabular}{ccc}
        \toprule
        U-Net size & AP \\
        \midrule
        0 & 25.8 \\
        1 & 37.9 \\
        2 & 45.4 \\
        3 & 46.3 \\
        \textbf{4} & 47.3 \\
        \bottomrule
    \end{tabular}
    \caption{Results of a study of the tiny U-Net size on the ScanNet v2 validation set. We use four feature levels by default.}
    \label{tab:ablation-unet}
\end{table}

\inline{Number of initial proposals.} Furthermore, we investigate the dependency between the number of initial proposals, accuracy, and runtime. Note that the number of proposals are approximate, since they cannot be set explicitly but manipulated through the NMS hyperparameters. 
Expectedly, the more proposals, the higher is AP (Tab.~\ref{tab:ablation-proposals}). However, with as many as 60 proposals, our method reaches the plateau in terms of segmentation accuracy. In the meantime, the inference time tends to increase with the growing number of proposals. Overall, we assume that with $\approx60$ initial proposals, our method demonstrates a decent trade-off between accuracy and speed, so we use this value by default in our experiments.

\begin{table}[ht!]
    \centering
    %\small
    \begin{tabular}{ccc}
    \toprule
    \#Initial & \multirow{2}{*}{AP} & Runtime \\
    proposals & & (in sec) \\
    \midrule
    $\approx$140 & 47.6 & 0.260 \\
    $\approx$100 & 47.5 & 0.210 \\
    $\approx$\textbf{60} & 47.3 & 0.140 \\
    $\approx$20 & 45.6 & 0.105 \\
    \bottomrule
    \end{tabular}
    \caption{Results of a study of the approximate number of initial proposals on the ScanNet v2 validation set. 60 object proposals are chosen as a default value, as it serves a good balance of accuracy and speed.}
    \label{tab:ablation-proposals}
\end{table}

\begin{table}[b!]
    \centering
    %\small
    \begin{tabular}{cccccc}
    \toprule
    Threshold & AP & AP\textsubscript{50} \\
    \midrule
    0.10 & 43.9 & 69.2 \\
    0.15 & 45.9 & 70.8 \\
    \textbf{0.20} & 47.3 & 71.2 \\
    0.25 & 47.9 & 70.9 \\
    0.30 & 48.2 & 70.6 \\
    0.40 & 48.1 & 70.1  \\
    \bottomrule
    \end{tabular}
    \caption{Results of TD3D with different point binary segmentation thresholds, obtained on the ScanNet v2 validation set. The threshold of 0.2 allows for the highest quality.}
    \label{tab:ablation-binary-thr}
\end{table}

\begin{table}[!h]
    \centering
    %\small
    \begin{tabular}{cccccc}
    \toprule
    FCAF3D assigner & IoU assigner & AP & AP\textsubscript{50} \\
    \midrule
    \checkmark &            & 46.3 & 70.2 \\
               & \checkmark & 45.5 & 69.6 \\
    \checkmark & \checkmark & \textbf{47.3} & \textbf{71.2} \\
    \bottomrule
    \end{tabular}
    \caption{Results of the proposal refinement model with different assigners. The FCAF3D assigner slightly outperforms the IoU assigner in a single-assigner mode, but the best scores are obtained with their combination. Accordingly, we use FCAF3D+IoU assigners.}
    \label{tab:ablation-two-assigners}
\end{table}

\begin{table}[!h]
    \centering
    %\small
    \begin{tabular}{cccccc}
    \toprule
    Threshold & AP & AP\textsubscript{50} \\
    \midrule
    \textbf{0.00} & 47.3 & 71.2 \\
    % 0.12 & 46.2 & 71.1 \\
    0.25 & 46.2 & 71.1 \\
    % 0.37 & 46.2 & 71.1 \\
    0.50 & 46.0 & 71.0 \\
    % 0.63 & 45.9 & 71.0  \\
    0.75 & 45.6 & 69.3  \\
    % 0.89 & 41.5 & 65.3  \\
    \bottomrule
    \end{tabular}
    \caption{Results of the IoU assigner with different IoU thresholds, obtained on the ScanNet v2 validation set. Thresholding with 0.0 provides the best results, so we assume that filtering is not needed.}
    \label{tab:ablation-iou-assigner-thr}
\end{table}

\inline{Point classification threshold.} We also vary the point binary segmentation threshold in 3D tiny U-Net, which is used to identify points either as \textit{foreground} or \textit{background} on inference. The results are presented in the Tab.~\ref{tab:ablation-binary-thr}. As can be observed from the Tab.~\ref{tab:ablation-binary-thr}, as the point binary segmentation threshold rises from 0.1 to 0.2, the AP\textsubscript{50} value experiences a noticeable increase, hitting the highest score of 71.2 at a threshold of 0.2. For the larger values, the AP\textsubscript{50} declines gradually, so the optimal value is defined unambiguously.

\inline{Assigners in the proposal refinement model.} Tab.~\ref{tab:ablation-two-assigners} presents the results of models trained using two different assigners: FCAF3D assigner and IoU assigner, individually and in combination. Taken individually, the FCAF3D assigner outperforms the IoU assigner, and the combination slightly improves the performance compared to using only one assigner, so we use the two of them by default. The Tab.~\ref{tab:ablation-iou-assigner-thr} shows the results obtained by varying IoU threshold value in the IoU assigner on the ScanNet v2 validation set. Evidently, filtering by threshold is redundant, since the highest AP and AP\textsubscript{50} values are obtained with the threshold of 0.0.

\inline{RoI extractor threshold.} The RoI extractor algorithm is parameterized with the minimum number of voxels in the proposal. If a proposal contains fewer voxels, it is discarded and not used further at the subsequent stages. The Tab.~\ref{tab:roi-extractor} shows the results of the RoI extractor algorithm with different minimum voxel thresholds on the ScanNet v2 validation set. As can be seen, when the minimum voxel threshold is between 1 and 200, both the AP and AP\textsubscript{50} scores remain constant at 47.3 and 71.2, respectively. However, when the threshold surpasses 200, its further increasing causes the degradation of the performance, so any number between 1 and 200 can be used as a default option.

\begin{table}[!h]
    \centering
    %\small
    \begin{tabular}{cccccc}
    \toprule
    Threshold & AP & AP\textsubscript{50} \\
    \midrule
    \textbf{1}    & 47.3 & 71.2 \\
    \textbf{10}   & 47.3 & 71.2 \\
    \textbf{50}   & 47.3 & 71.2 \\
    \textbf{100}  & 47.3 & 71.2 \\
    \textbf{200}  & 47.3 & 71.2 \\
    500  & 45.6 & 70.1  \\
    700  & 44.6 & 68.4  \\
    1000 & 42.6 & 65.1  \\
    \bottomrule
    \end{tabular}
    \caption{Results of the RoI extractor algorithm with different minimum voxel thresholds on the ScanNet v2 validation set. Any value between 1 and 200 can be used, since all of them ensure the same final quality.}
    \label{tab:roi-extractor}
\end{table}

\section{Conclusion}

In this work, we introduced TD3D, a novel 3D instance segmentation method following a top-down paradigm. Being fully-convolutional and trained end-to-end in a data-driven way, it does not rely on prior assumptions about the objects, which eases the burden of manually tuning domain-specific hyperparameters. We evaluated our method on the standard benchmarks: indoor ScanNet v2, ScanNet200, S3DIS, and aerial STPLS3D. Our experiments demonstrated that TD3D is on par with state-of-the-art grouping-based 3D instance segmentation methods: but being as accurate, it is more than $1.9$x faster on inference.

%%%%%%%%% REFERENCES
{\small
\bibliographystyle{ieee_fullname}
\bibliography{refs}
}

\end{document}